\documentclass[letterpaper, 10 pt, conference]{ieeeconf}  % Comment this line out if you need a4paper

\IEEEoverridecommandlockouts                              % This command is only needed if 
\usepackage{graphicx}
\usepackage{amsmath}
\usepackage{amssymb}
\usepackage{booktabs}
\usepackage{multirow}
\usepackage{booktabs}
\usepackage{graphicx}
\usepackage{paralist}
\usepackage[table]{xcolor}   % for row coloring
\usepackage{xcolor} % loads color support

\usepackage{newunicodechar}
\newunicodechar{−}{\ensuremath{-}}

\newcommand{\approach}{IMPRINT}

\usepackage[pagebackref,breaklinks,colorlinks]{hyperref}

\title{\LARGE \bf
\approach: Image-Conditioned Query Enrichment for Long-Tail Object Goal Navigation
}

\author{Jelin Raphael Akkara$^{1, 2}$, Filippo Ziliotto$^{1, 2}$, Luciano Serafini$^{2}$, Lamberto Ballan$^{1}$, Tommaso Campari$^{2}$%
\thanks{$^{1}$University of Padova, Italy. 
        {\tt\small 
        lamberto.ballan@unipd.it}}%
\thanks{$^{2}$Fondazione Bruno Kessler (FBK), Trento, Italy.
        {\tt\small \{jakkara, fziliotto, serafini, tcampari\}@fbk.eu}}%
}

\begin{document}

\maketitle
\thispagestyle{empty}
\pagestyle{empty}

%%%%%%%%%%%%%%%%%%%%%%%%%%%%%%%%%%%%%%%%%%%%%%%%%%%%%%%%%%%%%%%%%%%%%%%%%%%%%%%%

\begin{abstract}

Embodied AI increasingly relies on queryable semantic maps built from pre-trained vision–language models to enable zero-shot Object Goal Navigation (ObjectNav). However, existing approaches typically depend on text-only queries, which become less reliable as semantic specificity increases toward fine-grained object categories. We introduce \approach, a zero-shot plug-and-play framework that enriches textual object queries with web-sourced images to improve grounding in queryable maps. Retrieved images are encoded using a vision–language model, matched against the semantic map to produce similarity maps, and aggregated to yield context-aware localization. Notably, this requires no training or modification of the underlying navigation policy. To explicitly evaluate long-tail behavior, we present \textbf{HSSD-rare}, a new ObjectNav benchmark built on Habitat Synthetic Scenes and featuring semantically specific subcategories. Across both OVON and HSSD-rare, image-conditioned queries consistently improve object grounding and yield end-to-end navigation gains. Further analysis reveals that translating localization gains to navigation performance depends critically on downstream detection quality, highlighting a key systems bottleneck in long-tail embodied navigation. Code and dataset available at: \href{https://github.com/JelinR/IMPRINT}{github.io/IMPRINT}

% Further analysis reveals that while localization improves substantially, its translation to navigation performance depends critically on downstream detection quality, highlighting a key systems bottleneck in long-tail embodied navigation. Code and dataset available at: \href{https://github.com/JelinR/IMPRINT}{github.io/IMPRINT}

\end{abstract}

%%%%%%%%%%%%%%%%%%%%%%%%%%%%%%%%%%%%%%%%%%%%%%%%%%%%%%%%%%%%%%%%%%%%%%%%%%%%%%%%

\section{INTRODUCTION}

Recent advances in Embodied AI have enabled agents to interact intelligently with complex environments through semantic mapping and language grounding. A prominent approach is to construct queryable semantic maps by aligning textual queries with visual embeddings using pre-trained models such as CLIP~\cite{Radford2021}, BLIP-2~\cite{li2023blip}, and SED~\cite{xie2024sed}. These representations support zero-shot Object Goal Navigation (ObjectNav) by allowing agents to localize objects through open-set text queries without retraining~\cite{yokoyama2024vlfm, Busch2024}.

While effective for canonical object categories, text-only grounding becomes less reliable as semantic specificity increases. When queries refer to fine-grained or product-level subcategories, a category name alone may provide insufficient visual cues for accurate localization.

Meanwhile, web image search engines provide abundant visual examples for both common and rare object categories. Retrieval-based approaches in computer vision have leveraged web-sourced images to improve recognition of highly specific categories~\cite{sidhu2025search, jia2025visualwebinstruct}. However, such retrieval mechanisms have not been systematically integrated into embodied navigation pipelines, particularly as an accessible plug-and-play mechanism.

In this work, we introduce \approach, a plug-and-play framework for ObjectNav baselines built upon queryable semantic maps. Rather than modifying the underlying navigation policy, it enriches textual object queries with web-retrieved visual examples to improve target localization. Inspired by human search behavior, where users consult image examples to identify unfamiliar objects, \approach\ retrieves $N$ relevant images for a given query, encodes them using a vision–language model, matches each embedding against the semantic map, and aggregates the resulting similarity-scored maps to produce candidate object locations. The method is fully zero-shot, and model-agnostic, requiring no retraining.

We evaluate \approach\ across both canonical and fine-grained regimes. On OVON-syn~\cite{yokoyama2024hm3d}, which introduces increased semantic specificity over common categories, our method yields consistent end-to-end navigation gains. To explicitly study long-tail behavior, we introduce HSSD-rare, a new ObjectNav benchmark built upon Habitat Synthetic Scenes~\cite{khanna2024habitat}, featuring semantically specific subcategories derived from richly annotated 3D assets.

Beyond demonstrating improved semantic grounding in isolation, we analyze how these gains translate to active navigation. Our experiments show that while image-conditioned queries substantially improve semantic localization, their end-to-end impact depends critically on downstream pipeline components—particularly object detection. Controlled interventions reveal that improving detection quality unlocks additional navigation gains, highlighting a key bottleneck in long-tail embodied navigation.

\begin{figure}[!t]
    \centering
    \includegraphics[width=\columnwidth]{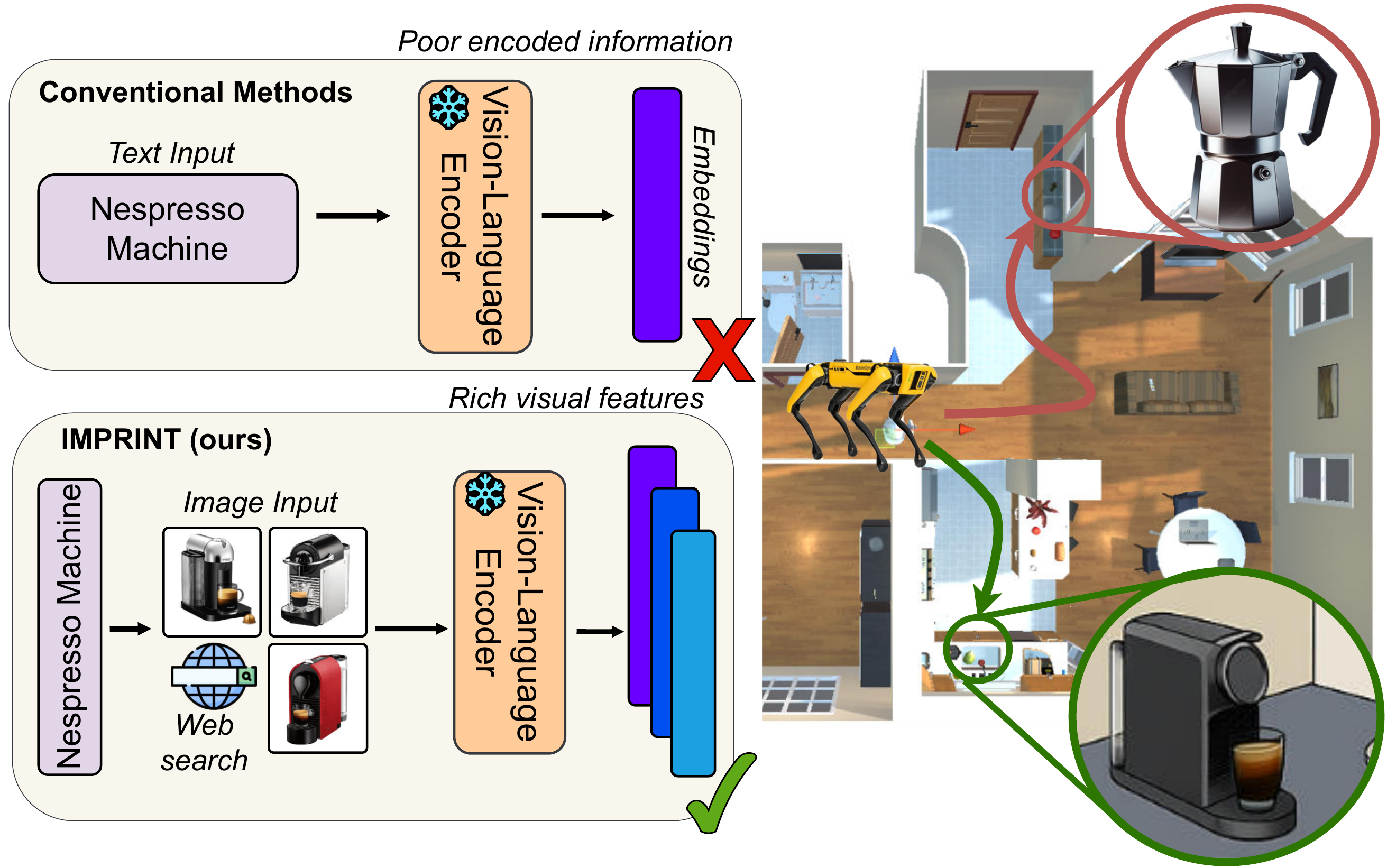}
    \caption{We present \approach, a simple yet effective plug-and-play method for retrieving open-set objects in a mapped environment. Given an object category, \approach\ queries the web for relevant images, extracts their embeddings, and leverages this information to conduct object localization.}
    \label{fig:teaser}
\vspace{-0.5cm}
\end{figure}

To sum up, our key contributions are as follows: \textit{(i)} We propose \approach, a plug-and-play framework that enhances open-vocabulary ObjectNav by enriching text queries with web-sourced visual context, improving grounding across both canonical and long-tail categories.
\textit{(ii)} We introduce HSSD-rare, a benchmark for fine-grained long-tail ObjectNav constructed from richly annotated synthetic scenes.
\textit{(iii)} We provide extensive zero-shot evaluations and ablations, revealing both the effectiveness of image-conditioned grounding and a systems-level bottleneck linking detection quality to end-to-end navigation performance.

%%%%%%%%%%%%%%%%%%%%%%%%%%%%%%%%%%%%%%%%%%%%%%%%%%%%%%%%%%%%%%%%%%%%%%%%%%%%%%%%

\section{Related Works}

% In this section, we review the prominent approaches in long-tailed object navigation.

In this section, we review prior work on open-vocabulary object navigation, semantic grounding with vision–language models, and long-tail recognition in embodied settings.

\noindent{\textbf{Open-Vocabulary Navigation and Grounding:}}
Early semantic mapping approaches relied on closed-vocabulary labels within photorealistic 3D datasets such as Matterport3D and Gibson~\cite{chang2017matterport3d, xia2018gibson}, limiting recognition to predefined categories. The introduction of vision–language models (VLMs) enabled open-vocabulary spatial representations~\cite{Huang2023, yokoyama2024vlfm, Busch2024}. VLMaps~\cite{Huang2023} and OpenFusion~\cite{Yamazaki2024} embedded CLIP features into 3D maps for free-text querying, while VLFM~\cite{yokoyama2024vlfm}, OVON~\cite{yokoyama2024hm3d}, and OneMap~\cite{Busch2024} adapted such queryable maps for zero-shot ObjectNav.

Our work builds on this paradigm by introducing image-conditioned query enrichment, improving grounding across both canonical categories and semantically specific long-tail subcategories. Unlike ImageNav or InstanceNav, which target specific visual instances or viewpoints, our setting remains category-level but requires fine-grained subcategory discrimination within open-vocabulary navigation.

On the other hand, Open-vocabulary grounding is driven by large-scale VLMs such as CLIP~\cite{Radford2021}, BLIP-2~\cite{li2023blip}, and SigLIP \cite{zhai2023sigmoid}, which align image and text embeddings through contrastive learning. Dense extensions including LSeg~\cite{Li2022b} and OpenScene~\cite{Peng2023} enable pixel- and 3D-level querying. Patch-level encoders such as SED~\cite{xie2024sed} further improve fine-grained localization.

Open-vocabulary detectors—including OwLv2~\cite{minderer2023scaling}, GroundingDINO \cite{liu2024grounding}, and YOLOWorld \cite{cheng2024yolo}—extend text conditioning to detection. While these models enable zero-shot localization, their integration into embodied pipelines introduces architectural sensitivities. Our approach complements this line of work by enriching textual queries with retrieved visual examples and revealing detector-dependent effects when translating object grounding improvements to navigation performance.

\noindent{\textbf{Long-Tail Embodied Settings:}}
Long-tail recognition addresses the imbalance between frequent and rare categories in large-scale visual data~\cite{deeplongtailedlearning}. Although VLM pretraining improves zero-shot generalization, fine-grained and commercially specific variants remain challenging, particularly under partial observability and spatial constraints in embodied environments. Most ObjectNav benchmarks emphasize broad semantic categories, even when new categories are introduced \cite{ramakrishnan2021habitat, yokoyama2024hm3d}. We instead focus on subcategory-level localization within realistic 3D scenes and analyze how long-tail semantics alter navigation failure regimes.

%%%%%%%%%%%%%%%%%%%%%%%%%%%%%%%%%%%%%%%%%%%%%%%%%%%%%%%%%%%%%%%%%%%%%%%%%%%%%%%%

\section{Problem Definition}
\label{sec: problem_def}

We consider the task of \textit{Object Goal Navigation} (ObjectNav), in which an embodied agent operating in a partially observable 3D environment must navigate to a target object category. The agent is provided with a natural language query $q$ specifying only the object category. At each time step $t$, it receives an egocentric observation
\[
o_t = (o_t^{\text{rgb}}, o_t^{\text{depth}}, p_t),
\]
where $o_t^{\text{rgb}}$ and $o_t^{\text{depth}}$ denote RGB and depth inputs, and $p_t$ represents the agent’s egocentric pose. Based on its observation history, the agent selects actions $a_t \in \mathcal{A}$ from a discrete navigation action space.

An episode terminates upon execution of the \textit{stop} action or after a maximum horizon of $T_{\max} = 500$ steps. Success is defined as the agent stopping within 1\,m of any instance of the queried object category.
The objective is to learn a policy $\pi(a_t \mid o_{\leq t}, q)$ that maximizes efficient and successful navigation under these constraints.

Beyond commonly studied object categories (e.g., \textit{chair}), we consider a fine-grained long-tail regime in which targets correspond to semantically specific subcategories of broader object classes (e.g., \textit{IKEA Markus chair}). 
We define long-tail categories as high-specificity variants that share visual and semantic overlap with a common parent category but require discriminating between closely related subcategories. 
Such targets are typically underrepresented in large-scale vision–language pretraining corpora and exhibit sparse instance-level coverage within scenes (Sec.~\ref{sec:dataset}).

%%%%%%%%%%%%%%%%%%%%%%%%%%%%%%%%%%%%%%%%%%%%%%%%%%%%%%%%%%%%%%%%%%%%%%%%%%%%%%%%

\section{Method}

\approach\ operates in two settings: \textit{Static} and \textit{Online}. In both, the agent retrieves web-sourced images for the target query, encodes them using a vision–language model, matches each embedding against the queryable map to produce similarity maps, and aggregates these maps to obtain the final target localization.

The phases differ in how the queryable map is constructed and evaluated. In the \textit{Static} phase, the map is pre-built and remains static, isolating the target localization component from downstream navigation factors. This controlled setting enables direct evaluation of grounding performance, without confounding effects from additional modules such as object detection or exploration policies.

In contrast, the \textit{Online} phase reflects the general embodied setting, where the agent incrementally constructs the queryable map online while navigating. Mapping and querying occur jointly, requiring the agent to balance exploration for map construction with goal-directed navigation.

\begin{figure*}[!t]
    \centering
    \includegraphics[width=1.\textwidth]{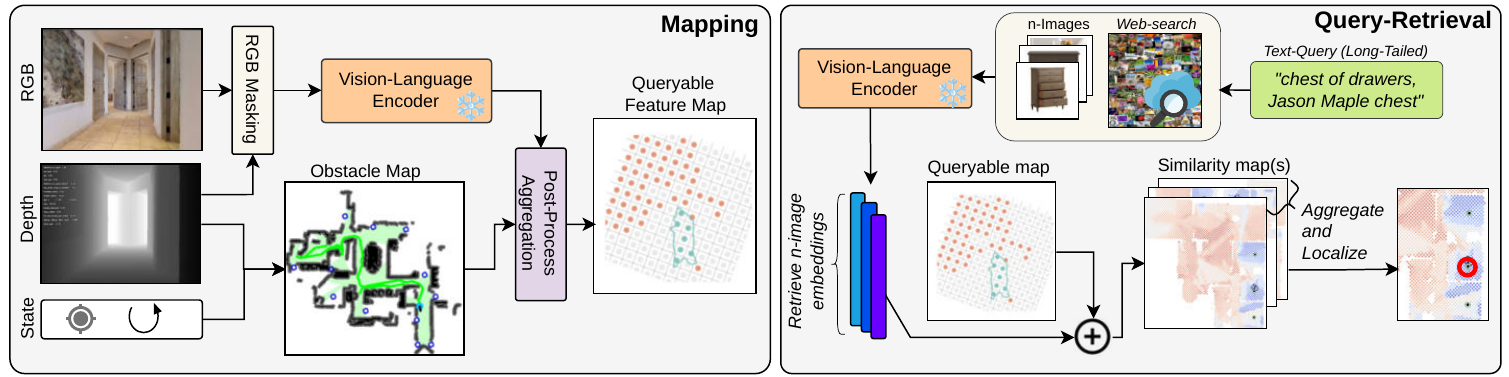}
    
    \caption{\textbf{\approach\ pipeline}. 
    (left) RGB observations are encoded by a vision–language model and projected into a queryable semantic feature map. 
    (right) Given a text query, relevant web images are retrieved and encoded; each embedding produces a similarity map over the stored feature map, and these maps are aggregated to yield a context-aware target localization. 
    In the static phase, mapping is completed first and object grounding is evaluated in isolation. In the online phase, mapping and grounding are performed jointly within the navigation loop.}
    \label{fig:pipeline}
\vspace{-0.5cm}
\end{figure*}

In the following sections, we detail these phases in more detail.

\subsection{Web-Image Retrieval}

Given a target text query, \approach\ enriches the query with relevant web-sourced images. We retrieve the top-$n$ candidate images, where $n$ is selected via ablation (Sec. \ref{sec:ablations}). For long-tail categories, we prepend the head category to improve retrieval quality (e.g. chair, Upholstered Drop Chair).

To ensure relevance, we apply a metadata gating step that retains images only if the head noun appears in either the ALT text or the URL path. Duplicate images are further removed. The resulting filtered set is then used for visual embedding.
Retrieved images are cached locally to avoid repeated downloads across episodes. In practice, retrieval introduces minimal overhead, with an average latency of $0.61 \pm 0.16$ seconds per image.

\subsection{Queryable Semantic Map}
\label{sub_sec: query_map}

\noindent{\textbf{Map Construction.}}
We construct a discrete global semantic map by projecting local visual semantics onto a 2D spatial grid, following the mapping pipeline of VLFM~\cite{yokoyama2024vlfm}.
At time $t$, the agent receives
\[
o_t^{\text{rgb}} \in \mathbb{R}^{H \times W \times 3}, 
\quad
o_t^{\text{depth}} \in \mathbb{R}^{H \times W}.
\]
The RGB observation is encoded using a vision-language model to obtain a semantic embedding $f_t \in \mathbb{R}^{d}$. The depth map is projected into a 3D point cloud using camera intrinsics and pose $p_t$, then discretized into a spatial grid aligned with a global map $M \in \mathbb{R}^{h \times w \times d}$, where $h,w$ denote grid resolution along the $x$ and $y$ axes and $d$ the embedding dimension.
For each visible grid cell $(i,j)$, the feature $f_t$ is written to $M_{i,j}$. Cells observed multiple times are aggregated via averaging, yielding a globally consistent semantic representation.

The above formulation assumes a holistic image embedding (e.g., BLIP2~\cite{li2023blip}). For patch-based  like SED ~\cite{xie2024sed}, we instead obtain a set of patch embeddings
\[
\{f_t^{(k)}\}_{k=1}^{K}, \quad f_t^{(k)} \in \mathbb{R}^{d},
\]
which are then projected via depth to their respective spatial footprints, enabling more fine-grained cell-wise semantic assignment.

\noindent{\textbf{Map Querying.}}
In a standard pipeline involving queryable maps, the target text query $q$ is encoded using a vision-language model to obtain a feature embedding $z_q \in \mathbb{R}^{d}$. This embedding is matched against the semantic map $M \in \mathbb{R}^{h \times w \times d}$ using cosine similarity, computed at each grid cell as $s_{i,j} = \cos(z_q, M_{i,j})$, where $(i,j)$ indexes spatial locations.
The resulting similarity-scored map is used to localize candidate object regions, with the highest-scoring cell selected as the primary point of interest.

\vspace{0.1cm}
\noindent{\textbf{Image-Conditioned Queries.}}
To further contextualize the text query, we retrieve $N$ relevant web images and encode them to obtain visual embeddings $\{z_q^{(k)}\}_{k=1}^{N}$, where $z_q^{(k)} \in \mathbb{R}^{d}$. Each embedding is independently matched against the semantic map, producing $N$ similarity maps computed as $s_{i,j}^{(k)} = \cos(z_q^{(k)}, M_{i,j})$.
These similarity maps are aggregated via averaging along the feature axis to obtain the final similarity map, i.e., $s_{i,j} = \frac{1}{N}\sum_{k=1}^{N} s_{i,j}^{(k)}$. Alternative aggregation strategies are explored in the ablation studies (Sec. \ref{sec:ablations}).

In settings where image-conditioned querying alone yields limited gains, we additionally incorporate the text embedding $z_q$ within the same aggregation framework. Concretely, the text-derived similarity map is combined with the image-derived maps during aggregation, enabling joint reasoning over textual semantics and visual examples.

\subsection{Static Phase: Isolated Object Grounding}

The Static phase aims to construct a scene-level queryable map that serves as an isolated test bed for evaluating target object localization, decoupled from downstream navigation dynamics.

To obtain full spatial coverage, we adapt a VLFM-based exploration pipeline in which the agent autonomously explores the environment using a frontier-based strategy. Exploration is performed in a target-free manner, allowing the agent to iteratively visit and map all reachable frontiers without predefined waypoints or goal supervision. Mapping is capped at 3000 steps to bound exploration time while ensuring sufficient environmental coverage.

Accurate semantic mapping requires reliable alignment between visual features and spatial projections. To prevent distant observations from introducing projection noise, we depth-mask the RGB input, retaining only pixels within a 4\,m radius of the agent and setting others to neutral values. This restricts feature extraction to geometrically reliable regions.

To further enforce dense mapping, we reduce the depth sensor’s maximum range to 4\,m. Consequently, the agent is forced to navigate closer to the frontier in order to update it, improving projection fidelity and spatial accuracy.

For patch-based feature extractors, projection reliability can be further improved by discarding patch embeddings associated with sharp depth discontinuities. This prevents geometrically unreliable projections and enables more reliable and finer-grained semantic assignment to grid cells.

The resulting scene-level semantic map is cached locally and used for static target localization evaluation without navigation (Fig \ref{fig:pipeline}). The predicted target location is defined as the highest-scoring cell in the map.

\subsection{Online Phase: ObjectGoal Navigation}

The Online phase evaluates whether improvements in target localization translate to end-to-end navigation performance in previously unseen environments under zero-shot conditions (Sec.~\ref{sec: problem_def}). Unlike the Static setting, the agent incrementally constructs and queries the semantic map at each time step, forming a coupled mapping–grounding–decision loop. Image-conditioned (optionally combined with text-conditioned) embeddings are used to produce similarity maps that guide online navigation (Fig \ref{fig:pipeline}).

In standard queryable-map pipelines \cite{yokoyama2024vlfm, Busch2024}, the agent greedily explores frontier regions with high similarity scores while continuously checking for candidate object detections. When a reliable detection is confirmed—often through consistency checks between detection outputs and high-scoring map regions—the agent transitions to goal-directed navigation using a PointGoal policy. If the detection is rejected, exploration continues.

\approach\ integrates into this loop as a plug-and-play module that only modifies semantic grounding. The navigation policy and final object-detection mechanism remain unchanged from the underlying baseline. As a result, any navigation improvements arise from better map-based points of interest produced by improved target grounding.

%%%%%%%%%%%%%%%%%%%%%%%%%%%%%%%%%%%%%%%%%%%%%%%%%%%%%%%%%%%%%%%%%%%%%%%%%%%%%%%%

\section{Dataset}
\label{sec:dataset}

\begin{figure}[!t]
    \centering
    \includegraphics[width=\columnwidth]{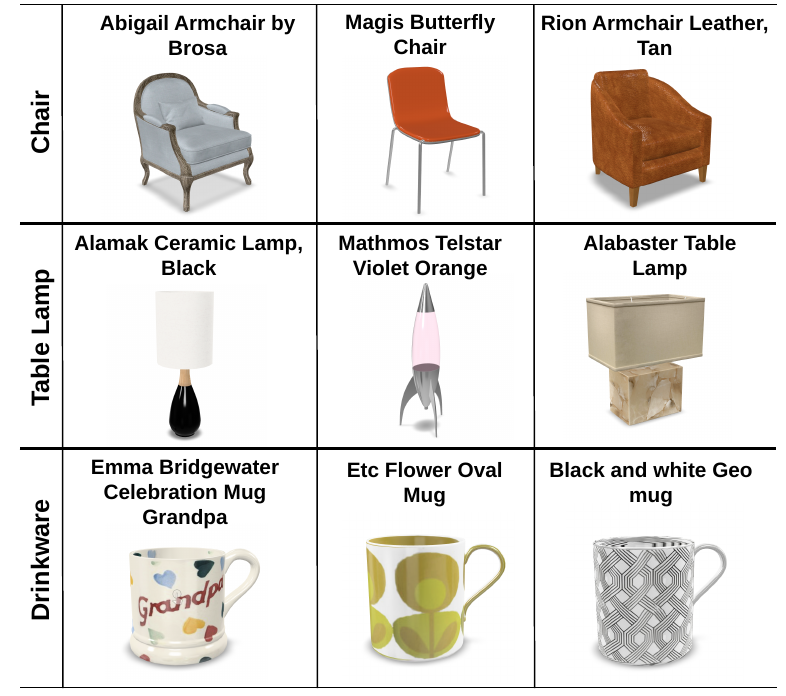}
    \caption{\textbf{HSSD-rare semantic categories}: Categories are organized hierarchically: each row corresponds to a main category, while columns illustrate long-tail subcategory variation.}
    \label{fig:longtail_imgs}
\vspace{-0.5cm}
\end{figure}

To evaluate object localization and active navigation performance in the long-tailed regime, we introduce \textit{HSSD-rare}, a benchmark designed to target fine-grained object categories. Here, long-tail refers to highly specific subcategories within broadly known object classes (Fig.~\ref{fig:longtail_imgs}).

We build on the Habitat Synthetic Scenes Dataset (HSSD), which comprises 211 high-quality synthetic 3D environments spanning 85 main object categories and 14,128 unique fine-grained subcategories. Each object instance is richly annotated with spatial position and geometric attributes. However, existing HSSD navigation episodes are limited to six coarse object classes and do not leverage the fine-grained subcategory annotations, leaving the vast majority of annotated categories unexplored. 

To address this limitation, we introduce HSSD-rare, a benchmark consisting of ObjectGoal navigation episodes where fine-grained subcategories serve as target goals. The dataset preparation procedure is described below.

\vspace{0.1cm}
\noindent{\textbf{Scene and Category Refinement.}}
To ensure reliable navigation and mapping, we refine both scene selection and object annotations. HSSD scenes span multiple floors and include indoor–outdoor regions, yet the dataset does not provide explicit floor assignments or connectivity annotations. This ambiguity can disrupt frontier-based exploration, as agents may inadvertently transition between floors or encounter inaccessible regions (e.g., closed doors).

We therefore restrict the benchmark to 17 scenes with navigational connectivity and estimate floor structure by sampling volumetric points along the vertical axis. Object instances are assigned to floors by thresholding their heights relative to the inferred floor levels, reducing cross-floor ambiguity. In scenes where indoor and outdoor regions are not mutually accessible, we partition them into independent environments and treat each as a separate map.

From the filtered annotations, we curate 20 main object categories based on subcategory diversity and co-occurrence within scenes, ensuring that evaluation reflects fine-grained sub-category grounding rather than coarse category localization.

\vspace{0.1cm}
\noindent{\textbf{Viewpoint Generation.}}
Habitat episodes require both object positions and valid agent viewpoints from which the target is visible. While HSSD provides object positions and dimensions, annotated viewpoints are not available. We therefore develop a scalable viewpoint sampling pipeline for active navigation settings.

% For each object instance, we sample candidate viewpoints radially around the object at uniformly spaced angles. From each sampled location, we project rays toward the object’s 3D boundary (computed from its position and axis-aligned dimensions) and evaluate visibility using depth observations. A candidate viewpoint is retained only if the projected rays do not intersect walls and are not significantly occluded by foreground geometry, ensuring clear line-of-sight visibility. This procedure has shown to be particularly robust in cluttered scenes, with empirical visualizations confirming the same (Fig \ref{fig:viewpoints}).

% For each object instance, we generate viewpoints by first estimating its local boundary. Radial rays are cast from the object center, and transitions from non-traversable to traversable space are recorded as boundary points. Candidate viewpoints are then uniformly sampled around this boundary. Each candidate is evaluated by orienting the agent toward the object and comparing the expected object distance with the depth observation at the projected object location. The candidate is retained only if no closer surface, such as a wall or other barrier, occludes the object, yielding localized viewpoints with clear line-of-sight visibility. Empirical visualizations show that this procedure remains robust in cluttered scenes (Fig.~\ref{fig:viewpoints}).

\begin{figure}[!th]
    \centering
    \includegraphics[width=\columnwidth]{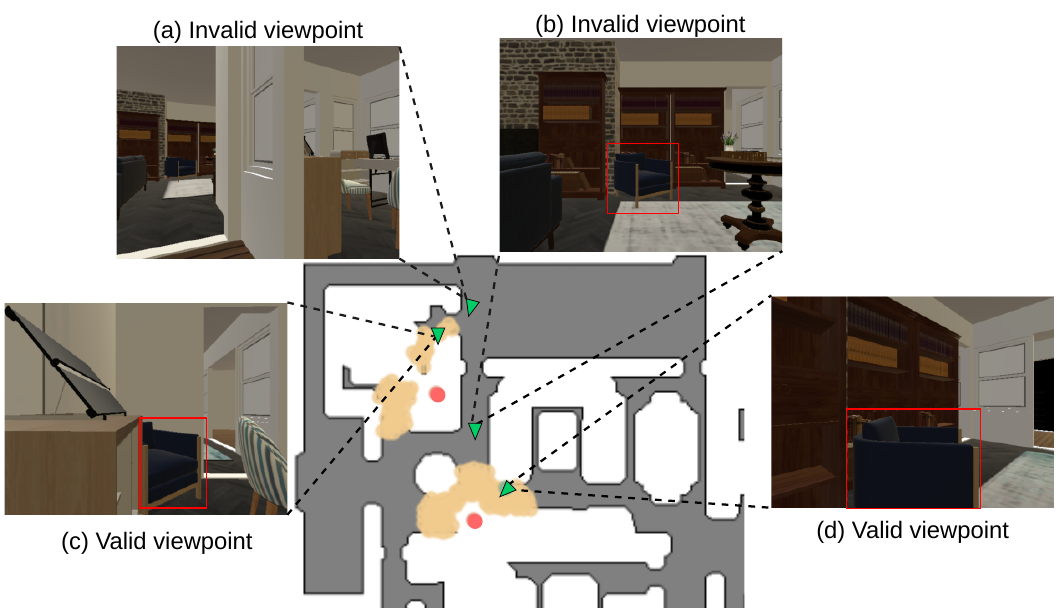}
    \caption{\textbf{Viewpoint generation for HSSD-rare.} 
    For each target instance (red), valid viewpoints (yellow) are identified by enforcing visibility and distance constraints. 
    Invalid examples (a,b) are rejected due to occlusion or excessive distance. 
    Valid viewpoints (c,d) demonstrate spatial diversity across cluttered and open regions, preventing trivial start–goal configurations.}
    \label{fig:viewpoints}
\vspace{-0.5cm}
\end{figure}

For each object instance, we generate viewpoints by first estimating its local boundary. Radial rays are cast from the object center, and transitions from non-traversable to traversable space are recorded as boundary points. Candidate viewpoints are then uniformly sampled around this boundary and oriented toward the object. Each candidate is retained only if the expected object distance is consistent with the observed depth at the projected object location, indicating that no closer surface occludes the object. This yields localized viewpoints with clear line-of-sight visibility, while also remaining robust in cluttered scenes (Fig.~\ref{fig:viewpoints}).

% \begin{figure}[!t]
%     \centering
%     % \includegraphics[width=\columnwidth]{IROS26/images/imprint_ablation_img.png} % Adjust width as needed
%     \includegraphics[width=\columnwidth]{IMPRINT/images/longtail_imgs.pdf}
%     \caption{\textbf{HSSD-rare semantic categories}: Categories are organized hierarchically: each row corresponds to a main category, while columns illustrate long-tail subcategory variation.}
%     \label{fig:longtail_imgs}
% \vspace{-0.5cm}
% \end{figure}

\vspace{0.1cm}
\noindent{\textbf{Episode Sampling.}}
For each main category, we generate a fixed budget of 50 episodes to ensure balanced evaluation across high-level object classes. Within each main category, we adopt a hierarchical sampling strategy designed to promote fine-grained subcategory discrimination. Let $\mathcal{S}$ denote the set of scenes, $\mathcal{C}_s$ the set of subcategories of the main category present in scene $s$, and $\mathcal{I}_{s,c}$ the set of instances belonging to subcategory $c$ in scene $s$. 

We first sample a scene $s \sim \mathrm{Cat}(p)$ with $p(s) \propto |\mathcal{C}_s|$, favoring scenes that contain a larger number of distinct subcategories of the selected main category. We then sample a subcategory $c \sim \mathrm{Unif}(\mathcal{C}_s)$, followed by an instance $i \sim \mathrm{Unif}(\mathcal{I}_{s,c})$. This strategy ensures balanced main-category evaluation while encouraging intra-class fine-grained discrimination among co-occurring subcategories.

\vspace{0.1cm}
\noindent{\textbf{Dataset Statistics.}}
The benchmark contains 1000 navigation episodes (50 per 20 main categories) across 17 scenes, with navigable area averaging 103.15 $\pm$ 51.09\,m$^2$. Start–goal geodesic distance averages 6.55 $\pm$ 4.31\,m, and each instance provides 83.22 $\pm$ 51.00 valid viewpoints, reflecting diverse trajectory lengths and target spatial configurations.

The dataset comprises 559 distinct fine-grained subcategories and 992 total instances, with multiple subcategories of the same main category frequently co-occurring within a scene. This setup requires the agent to distinguish between closely related subcategories rather than merely locating any instance of a broader class.

%%%%%%%%%%%%%%%%%%%%%%%%%%%%%%%%%%%%%%%%%%%%%%%%%%%%%%%%%%%%%%%%%%%%%%%%%%%%%%%%

\section{Experiments}
\label{sec: experiments}

% \approach\ provides a flexible framework that integrates seamlessly with existing embodied navigation pipelines. We evaluate our method on two open-vocabulary benchmarks featuring diverse object categories, enabling comparisons in both isolated object grounding and ObjectGoal Navigation. We further conduct extensive ablation studies to quantify the impact of key design choices.

We evaluate \approach\ on two open-vocabulary ObjectNav benchmarks, considering both isolated object grounding and end-to-end navigation. We further conduct ablations to quantify the impact of key design choices.

\subsection{Datasets and Metrics}

We evaluate \approach\ on two open-vocabulary ObjectNav benchmarks: OVON~\cite{yokoyama2024hm3d} and HSSD-rare, which capture complementary aspects of semantic specificity.

OVON provides three validation splits (\textit{val\_seen}, \textit{val\_unseen}, and \textit{val\_seen\_synonyms}). As our approach operates in a zero-shot setting, we focus on semantic granularity rather than seen/unseen distinctions. The \textit{val\_seen} split contains canonical household categories (e.g., chair, table), \textit{val\_unseen} introduces structurally diverse objects (e.g., statue, rug), and \textit{val\_seen\_synonyms} employs fine-grained variants of common classes (e.g., computer desk, dining chair). We evaluate primarily on \textit{val\_seen\_synonyms}, which provides increased semantic specificity while retaining visually common instances. Episodes are balanced across categories to avoid evaluation bias.

HSSD-rare extends HSSD-Hab~\cite{khanna2024habitat} by constructing ObjectNav episodes over fine-grained subcategories. Unlike OVON, which varies lexical specificity, HSSD-rare introduces true subcategory-level targets derived from richly annotated 3D assets, requiring intra-class discrimination among co-occurring variants. 

Together, these benchmarks enable evaluation across synonym-based semantic variation (OVON) and fine-grained subcategory localization (HSSD-rare).

% \paragraph{\textbf{Metrics}}
Regarding metrics, we evaluate both phases (static and online) using standard ObjectGoal Navigation metrics. Success Rate (SR) measures whether the agent successfully localizes the target within a 1\,m threshold. Success weighted by Path Length (SPL) evaluates navigation efficiency by comparing the agent's trajectory length to the shortest-path distance, conditioned on successful episodes. Distance to Goal (DTG) measures the geodesic distance from the agent’s final position to the target object. 
For the static phase (isolated object grounding) evaluation, we report SR and DTG only, as SPL requires an executed navigation trajectory.

\subsection{Baselines}

\noindent{\textbf{Static Phase Baselines.}}
We evaluate object grounding performance across widely used vision-language feature encoders. We consider BLIP2~\cite{li2023blip} and SigLIP~\cite{zhai2023sigmoid}, representing established and emerging encoders in embodied navigation pipelines. We further include patch-based encoders, with SED~\cite{xie2024sed} serving as a strong baseline for fine-grained semantic localization.

\vspace{0.1cm}
\noindent{\textbf{Online Phase Baselines.}}
We evaluate three zero-shot navigation pipelines with differing policy structures. VLFM~\cite{yokoyama2024vlfm} and OneMap~\cite{Busch2024} construct queryable semantic maps and employ frontier-based exploration guided by similarity scores. When grounding and object detection are validated, the agent transitions to goal-directed navigation using a PointGoal controller.

In contrast, ZSON~\cite{majumdar2022zson} conditions a pretrained policy network directly on the semantic query embedding without constructing an explicit spatial semantic map. This allows us to assess how query enrichment behaves under both map-based and direct policy-conditioning paradigms.

These systems also provide encoder diversity in the active setting: ZSON uses CLIP, VLFM employs BLIP2, and OneMap utilizes SED, enabling comparison across both policy design and feature representation.

\subsection{Results}
We report quantitative results across both the static and online phases below, followed by a failure analysis at the end. 

\vspace{0.1cm}
{\textbf{Static Phase (Object Grounding).}}
    % Passive Eval: All
    \setlength{\tabcolsep}{3pt} 
    \begin{table}[!t] 
    \caption{\scriptsize Results for static phase on OVON-syn~\cite{yokoyama2024hm3d} and HSSD-rare datasets, across three different feature encoders. Query modes vary as text-only, image-only, and a combination of text and image (Both).}
    \small 
    \centering 
    \begin{tabular}{lllcc} 
     \toprule 
    & \textbf{Encoder} & \textbf{Mode} & \textbf{SR} (1m)$\uparrow$ & \textbf{DTG} (m)$\downarrow$ \\ 
    
    \midrule 
    
    \multirow{9}{*}{\rotatebox{90}{\textbf{OVON-syn}}} &
        
        \multirow{3}{*}{\textbf{BLIP2}~\cite{li2023blip}}   
            & Text  & 38.71 & 3.27 \\ 
            & & Image & 42.26 & 2.89 \\ 
            & & Both  & \textbf{42.58} & \textbf{2.83} \\ 
        
        \cmidrule(lr){2-5} & 
        
        \multirow{3}{*}{\textbf{SigLIP}~\cite{zhai2023sigmoid}}
            & Text  & 7.74 & 5.74 \\ 
            & & Image & 28.39 & 3.30 \\ 
            & & Both  & \textbf{29.35} & \textbf{3.23} \\ 
        
        \cmidrule(lr){2-5} & 
        
        \multirow{3}{*}{\textbf{SED}~\cite{xie2024sed}}
            & Text  & 40.00 & 3.06 \\ 
            & & Image & \textbf{47.42} & \textbf{2.71} \\ 
            & & Both  & 44.52 & 2.76 \\ 
    
    \midrule \midrule 
    
    \multirow{9}{*}{\rotatebox{90}{\textbf{HSSD-rare}}} &
        
        \multirow{3}{*}{\textbf{BLIP2}~\cite{li2023blip}}
            & Text  & 17.11 & 6.96 \\ 
            & & Image & 19.69 & \textbf{6.48} \\ 
            & & Both  & \textbf{20.00} & 6.51 \\ 
        
        \cmidrule(lr){2-5} & 
        
        \multirow{3}{*}{\textbf{SigLIP}~\cite{zhai2023sigmoid}}
            & Text  & 2.15 & 9.50 \\ 
            & & Image & 16.51 & 6.96 \\ 
            & & Both  & \textbf{16.72} & \textbf{6.94} \\ 
        
        \cmidrule(lr){2-5} & 
        
        \multirow{3}{*}{\textbf{SED}~\cite{xie2024sed}} 
            & Text  & 23.87 & 6.07 \\ 
            & & Image & 26.05 & 5.61 \\ 
            & & Both  & \textbf{26.46} & \textbf{5.52} \\ 
    
    \bottomrule 
    \end{tabular} 
    \label{tab:passive_results} 
    \vspace{-0.5cm}
    \end{table}
Table~\ref{tab:passive_results} reports isolated object grounding performance, measuring the top-ranked prediction against the ground-truth target location. Across all encoders and datasets, conditioning queries on retrieved images consistently improves grounding performance. On OVON-syn, image-conditioned queries yield an average gain of $+10.54$ SR ($−1.05$\,m DTG) over text-only queries. Improvements remain substantial on HSSD-rare, with $+6.37$ SR ($−1.19$\,m DTG) gains, indicating that visual query grounding remains beneficial even under fine-grained long-tail conditions. Notably, performance between image-only and fused modes is closely matched, suggesting that retrieved visual context provides the dominant grounding signal.

% We evaluate isolated object grounding performance in Table~\ref{tab:passive_results}, measuring localization accuracy considering the top-ranked prediction and the ground-truth position.

% For each configuration, we consider the top-ranked prediction and measure localization accuracy against the ground-truth target position.
    
% We evaluate isolated object grounding performance after completing web-image retrieval and environment mapping. Table~\ref{tab:passive_results} presents a comparative analysis across text-only, image-conditioned, and fused (both) query modes, evaluated over three encoder types and two datasets. For each configuration, we consider the top-ranked prediction and measure localization accuracy against the ground-truth target position.

% Across all encoders and datasets, conditioning queries on retrieved images consistently improves grounding performance. On OVON-syn, image-conditioned queries yield an average gain of $+10.54$ SR ($−1.05$\,m DTG) over text-only queries. Improvements remain substantial on HSSD-rare, with $+6.37$ SR ($−1.19$\,m DTG) gains, indicating that visual query grounding remains beneficial even under fine-grained long-tail conditions. Notably, performance between image-only and fused modes is closely matched, suggesting that retrieved visual context provides the dominant grounding signal.

Comparing datasets, grounding performance is uniformly lower on HSSD-rare than OVON-syn across all encoders and query modes. This highlights the increased difficulty of localizing truly fine-grained and commercially specific objects, as opposed to synonym-based lexical variants of common categories.

Encoder-wise trends reveal additional insights. SED achieves the strongest localization performance on OVON-syn, reflecting the advantages of patch-based semantic encoding for spatial grounding. However, this advantage diminishes on HSSD-rare, where gains over text-only grounding are comparatively smaller. In contrast, holistic encoders such as BLIP2 and SigLIP maintain more consistent improvements across both datasets, suggesting greater robustness to shifts in fine-grained semantic distributions.

Overall, all configurations demonstrate improved SR and DTG over text-only grounding, across both common and long-tail object settings. The performance gap observed on HSSD-rare further underscores the challenges posed by long-tailed localization and highlights significant headroom for future progress in long-tail embodied grounding.

% Active Eval: All
\setlength{\tabcolsep}{4.2pt} 
\begin{table}[!t] 
\caption{\scriptsize Results for online phase across the OVON-syn~\cite{yokoyama2024hm3d} and HSSD-rare datasets. We compare three baselines, w and w/o \approach, which integrates image-queries into the standard pipeline.}
\small 
\centering 
\begin{tabular}{lllccc} 
\toprule 

& \textbf{Method} & \textbf{Mode} & \textbf{SR} (1m) $\uparrow$ & \textbf{SPL} $\uparrow$ & \textbf{DTG} (m)$\downarrow$ \\ 

\midrule 

\multirow{6}{*}{\rotatebox{90}{\textbf{OVON-syn}}} &
    
    % \multirow{2}{*}{\textbf{ZSON}~\cite{majumdar2022zson}}   
    \multirow{2}{*}{\textbf{ZSON}~\cite{majumdar2022zson}}   
          & Standard  & 20.33 & 10.51 & \textbf{4.44} \\
          & & IMPRINT & \textbf{21.33} & \textbf{10.66} & 4.61 \\
    
    \cmidrule(lr){2-6} & 
    
    % \multirow{2}{*}{\textbf{VLFM}~\cite{li2023blip}} 
    \multirow{2}{*}{\textbf{VLFM}~\cite{yokoyama2024vlfm}} 
          & Standard  & 23.83 & 10.63 & 4.41 \\
          & & IMPRINT & \textbf{27.17} & \textbf{12.16} & \textbf{4.17} \\
    
    \cmidrule(lr){2-6} & 
    
    % \multirow{2}{*}{\textbf{OneMap}~\cite{Busch2024}} 
    \multirow{2}{*}{\textbf{OneMap}~\cite{Busch2024}} 
          & Standard  & 11.33 & 4.98 & 4.84 \\
          & & IMPRINT & \textbf{13.00} & \textbf{5.32} & \textbf{4.53} \\

\midrule \midrule 

\multirow{6}{*}{\rotatebox{90}{\textbf{HSSD-rare}}} &
    
    % \multirow{2}{*}{\textbf{ZSON}~\cite{majumdar2022zson}}   
    \multirow{2}{*}{\textbf{ZSON}~\cite{majumdar2022zson}}   
          & Standard  & 5.99 & 2.00 & 6.51 \\
          & & IMPRINT & \textbf{6.46} & \textbf{2.67} & \textbf{6.49} \\
    
    \cmidrule(lr){2-6} & 
    
    % \multirow{2}{*}{\textbf{VLFM}~\cite{li2023blip}}
    \multirow{2}{*}{\textbf{VLFM}~\cite{yokoyama2024vlfm}}
          & Standard  & 12.74 & 4.94 & 6.40 \\
          & & IMPRINT & \textbf{13.91} & \textbf{5.46} & \textbf{5.74} \\
    
    \cmidrule(lr){2-6} & 
    
    % \multirow{2}{*}{\textbf{OneMap}~\cite{Busch2024}}  
    \multirow{2}{*}{\textbf{OneMap}~\cite{Busch2024}}  
          & Standard  & 6.58 & 3.18 & 6.25 \\
          & & IMPRINT & \textbf{7.05} & \textbf{3.65} & \textbf{5.91} \\

\bottomrule 
\end{tabular} 
% \caption{Active Navigation results.} 
\label{tab:online_eval}
\vspace{-0.5cm}
\end{table}

\noindent{\textbf{Online Phase (ObjectGoal Navigation).}}
We evaluate whether the static localization gains transfer to end-to-end ObjectGoal Navigation. Unlike isolated grounding, navigation performance depends not only on semantic localization but also on map projection fidelity, object detection accuracy, thresholding policies, and navigation control. Table~\ref{tab:online_eval} compares each baseline with our image-conditioned extension across both datasets.

% Table~\ref{tab:online_eval} reports results across two datasets and three navigation baselines, comparing the standard pipeline with our image-conditioned extension.

Across baselines, stronger feature encoders yield higher absolute performance (VLFM with BLIP2, OneMap with SED, ZSON with CLIP). ZSON underperforms VLFM and OneMap despite using a capable encoder, consistent with its architecture: it conditions a policy network directly on the query embedding, whereas VLFM and OneMap explicitly leverage similarity maps to trigger goal-directed navigation. As a result, improvements in localization are more directly expressed in the map-based pipelines.

\approach\ yields clear gains on OVON-syn, demonstrating reliable end-to-end benefits in the general-category regime. On HSSD-rare, gains are more modest, suggesting that long-tail navigation is more strongly constrained by downstream components. Given that the static results confirm improved long-tail grounding, this indicates that the bottleneck lies in how localization signals propagate through the navigation pipeline rather than in semantic grounding itself.

Failure analysis (Sec. \ref{para: failure_modes}) identifies mis-detection as a dominant error mode under long-tail conditions. To test whether detection quality limits performance, we enhance the detection stage in OneMap by conditioning YOLOWorld with the top-$3$ retrieved reference images. This modification yields substantial gains on both OVON-syn and HSSD-rare (Table~\ref{tab:online_eval_onemap_improved}), confirming that object detection quality constrains how semantic localization improvements translate into navigation success.

Category-wise analysis shows that gains are broadly distributed across fine-grained targets. The largest improvements occur in semantically specific categories such as \textit{treadmill}, \textit{clothing}, \textit{bottle}, and \textit{table lamp}. For other categories (e.g., \textit{shoes}, \textit{teapot}), performance remains comparable to the baseline, indicating that image-conditioned enrichment yields targeted improvements without degrading existing behavior.

Overall, \approach\ is effective across both regimes: it delivers strong improvements in general-category navigation and reveals additional extractable potential in long-tail settings when paired with sufficiently robust downstream components.

    % Active Eval: All
    \setlength{\tabcolsep}{4pt} 
    \begin{table}[!t] 
    \caption{\scriptsize Results for online phase with enhanced object detector module (IMPRINT*). We consider OneMap, and enhance its detector module, YoloWorld.}
    \small 
    \centering 
    \begin{tabular}{llccc} 
    \toprule 

    \textbf{Dataset} & \textbf{Mode} & \textbf{SR} (1m)$\uparrow$ & \textbf{SPL} $\uparrow$ & \textbf{DTG} (m)$\downarrow$ \\ 
    
    \midrule 
    
    % \multirow{6}{*}{\rotatebox{90}{\textbf{OVON-syn}}} &
        
        \multirow{3}{*}{\textbf{OVON-syn}}
              & Standard  & 11.33 & 4.98 & 4.84 \\
              & IMPRINT & 13.00 & 5.32 & \textbf{4.53} \\
              & IMPRINT* & \textbf{18.17} & \textbf{11.45} & 4.98 \\

    \midrule

        \multirow{3}{*}{\textbf{HSSD-rare}}   
              & Standard  & 6.58 & 3.18 & 6.25 \\
              & IMPRINT & 7.05 & 3.65 & \textbf{5.91} \\
              & IMPRINT* & \textbf{11.75} & \textbf{5.91} & 5.96 \\

    \bottomrule 
    \end{tabular}  
    \label{tab:online_eval_onemap_improved}
    \vspace{-0.5cm}
    \end{table}

\vspace{0.1cm}
\noindent{\textbf{Failure Analysis.}}
\label{para: failure_modes}
We analyze online navigation failures across four categories: \textit{Timeout}, \textit{Mis-detection}, \textit{Exhausted Exploration}, and \textit{Stuck}. Timeout dominates (55.3\%), yet only 4\% of these episodes enter the 2\,m target vicinity, indicating large-scale semantic misses rather than near-goal control errors. Mis-detection accounts for 29.8\% of failures, while exploration and stagnation are negligible. These patterns point to semantic localization and detection as primary constraints under long-tail conditions.

Crucially, enhancing the detection module yields substantial gains in both HSSD-rare and OVON-syn (Table \ref{tab:online_eval_onemap_improved}), indicating that object detection quality acts as a shared systems-level bottleneck that limits how improved grounding translates into navigation success.

\subsection{Ablations}
\label{sec:ablations}

% We conduct ablation studies on \approach\ at multiple levels. 

% Passive Eval: OVON
\setlength{\tabcolsep}{3pt} 
\begin{table}[!t] 
\caption{\scriptsize Object Grounding results on the OVON dataset splits~\cite{yokoyama2024hm3d}. We compare the performance of \approach\ (with BLIP2 extractor) across different query modes.}
\small 
\centering 
\begin{tabular}{llcc} 
\toprule 

\textbf{Split} & \textbf{Mode} & \textbf{SR} (1m)$\uparrow$ & \textbf{DTG} (m)$\downarrow$ \\ 

\midrule 

\multirow{3}{*}{\textbf{OVON-seen}}    
    & Text & 40.49 & 3.00 \\ 
    & Image & 41.03 & \textbf{2.73} \\ 
    & Both & \textbf{41.50} & 2.79 \\ 

\cmidrule(lr){1-4} 

\multirow{3}{*}{\textbf{OVON-synon.}} 
    & Text & 38.71 & 3.27  \\ 
    & Image & 42.26 & 2.89 \\ 
    & Both & \textbf{42.58} & \textbf{2.83} \\ 

\cmidrule(lr){1-4}

\multirow{3}{*}{\textbf{OVON-unseen}} 
    & Text & 31.19 & 3.43 \\ 
    & Image & \textbf{34.58} & 3.17 \\ 
    & Both & \textbf{34.58} & \textbf{3.15} \\

\bottomrule 
\end{tabular} 
\label{tab:ovon_static_results} 
% \vspace{-0.5cm}
\end{table}

\noindent{\textbf{Object Grounding in OVON.}}  
Table \ref{tab:ovon_static_results} reports isolated object grounding performance (BLIP2 encoder) using the three OVON splits—\textit{seen}, \textit{unseen}, and \textit{synonyms}—under text-only, image-only, and text+image query modes. 

As semantic granularity increases from the seen to the synonyms split, the gains from image-conditioned queries become more pronounced. Improvements are modest for seen categories (+1.01 SR), larger for unseen (+3.39 SR), and largest for synonyms (+3.87 SR), indicating that visual query enrichment is particularly beneficial under fine-grained lexical variation. Distance-to-goal (DTG) consistently decreases when incorporating images, confirming improved spatial localization. Notably, image-only and text+image modes perform comparably, suggesting that retrieved visual context provides the dominant grounding signal once incorporated.

\vspace{0.1cm}
\noindent{\textbf{Aggregation Strategy.}}
When multiple queries are used (e.g., multiple retrieved images or text+image prompts), the semantic map produces multiple similarity maps that must be aggregated into a final prediction. Table \ref{tab:method_ablation} compares three aggregation strategies: arithmetic mean, harmonic mean, and a hybrid scheme (arithmetic mean over text maps combined with harmonic mean over image maps). Across encoders and datasets, arithmetic averaging consistently yields the best performance, indicating that simple linear aggregation preserves useful similarity structure more effectively than harmonic or hybrid schemes.

    % Ablation: Mean
    \setlength{\tabcolsep}{4pt}
    \begin{table}[!t]
    \caption{\scriptsize Ablation of similarity-map aggregation strategies. Hybrid stands for a mixture of both simple mean and harmonic mean.}
    \centering
    \small
    \begin{tabular}{l l c c c}
    \toprule
     & & \multicolumn{3}{c}{\textbf{SR} (1m)$\uparrow$ } \\
    \cmidrule(lr){3-5}
     & & & \textbf{Method} & \\
    \cmidrule(lr){3-5}
    Encoder & Dataset & Mean & Harmonic mean & Hybrid \\
    
    \multirow{2}{*}{\textbf{BLIP2}~\cite{li2023blip}}
      & OVON-syn  & \textbf{42.58} & 19.35 & 40.00 \\
      & HSSD-rare & \textbf{20.00} & 11.18 & 18.26 \\
      
    \midrule
    
    \multirow{2}{*}{\textbf{SigLIP}~\cite{zhai2023sigmoid}}
      & OVON-syn & \textbf{29.35} & 4.19 & 8.71 \\
      & HSSD-rare & \textbf{16.72} & 1.23 & 1.74 \\
    
    \midrule
    
    \multirow{2}{*}{\textbf{SED}~\cite{xie2024sed}}
      & OVON-syn  & \textbf{44.52} & 4.84 & 41.61 \\
      & HSSD-rare & \textbf{26.46} & 2.46 & 24.41 \\
    \bottomrule
    \end{tabular}

    \label{tab:method_ablation}
    \vspace{-0.5cm}
    \end{table}

\begin{figure}[!t]
    \centering
    \includegraphics[width=\columnwidth]{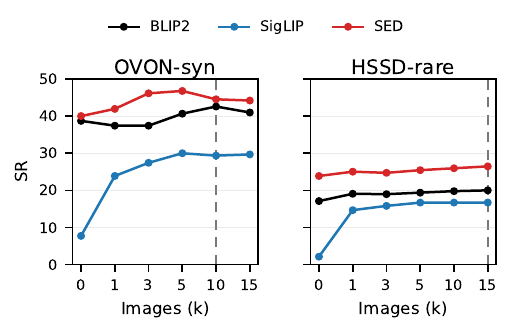}
    \caption{\textbf{Ablation with Number of Images}: Object grounding success rate is measured as the number of images (k) is varied, keeping the feature encoder fixed as BLIP2. Optimal values are marked as vertical lines.}
    \label{fig:ablation_imgs}
\vspace{-0.5cm}
\end{figure}

\vspace{0.1cm}
\noindent{\textbf{Number of Retrieved Images.}}
We further vary the number of retrieved web-images across feature encoders and datasets (Fig. \ref{fig:ablation_imgs}). Performance improves steadily with additional images up to a peak, after which gains saturate or slightly decline. OVON-syn peaks at $n=10$, while HSSD-rare peaks at $n=15$, suggesting that fine-grained long-tail categories benefit from more visual evidence. Beyond this range, additional images provide diminishing returns and may introduce noise.

% On OVON-syn, performance peaks at $n=10$ images, whereas on HSSD-rare the optimal value shifts higher at $n=15$. This suggests that semantically fine-grained long-tail categories benefit from additional visual evidence, while more generic categories require fewer reference images for effective grounding. Notably, excessive images do not consistently improve performance, indicating diminishing returns beyond the optimal range. Similar trends are noticed when using the other feature encoders.

\subsection{Limitations}
\label{sec:limitations}

% \approach\ relies on the availability of relevant web images for long-tailed object categories, as poor retrievals may introduce downstream errors. Moreover, evaluation on HSSD-rare, a synthetic benchmark, may not fully reflect real-world long-tailed object-goal navigation, motivating validation in more realistic benchmarks and real-world settings.

% \approach\ relies on the availability of relevant web images for long-tailed object categories, as poor retrievals may introduce downstream errors. Localization gains can also be constrained by navigation bottlenecks like detection module. Moreover, evaluation on HSSD-rare may not fully reflect real-world long-tailed object-goal navigation, motivating validation in more realistic settings.

% \approach\ relies on the availability of relevant web images for long-tailed object categories, and localization gains can also be constrained by navigation bottlenecks like the detection module. Moreover, evaluation on HSSD-rare may not fully reflect real-world long-tailed object-goal navigation, motivating validation in more realistic settings.

\approach\ relies on the availability of relevant web images for long-tailed object categories, and its localization gains can be constrained by pipeline bottlenecks such as object detection. Moreover, evaluation on HSSD-rare may not fully reflect real-world long-tailed ObjectGoal Navigation, motivating validation in more realistic settings.

% \approach\ relies on relevant web images for long-tailed object categories, as poor retrievals may introduce downstream errors. End-to-end gains can also be limited by pipeline bottlenecks such as object detection and policy execution. Finally, evaluation on HSSD-rare may not fully reflect real-world long-tailed ObjectGoal Navigation, motivating validation in more realistic settings. Finally, evaluation on synthetic HSSD-rare may not fully reflect real-world long-tailed ObjectGoal Navigation.

%%%%%%%%%%%%%%%%%%%%%%%%%%%%%%%%%%%%%%%%%%%%%%%%%%%%%%%%%%%%%%%%%%%%%%%%%%%%%%%%

\section{Conclusion}

We introduced \approach, a zero-shot framework that enriches text-based object queries with web-sourced visual context for queryable map grounding in embodied navigation. Across multiple encoders and datasets, \approach\ improves semantic localization in both common and fine-grained settings, with larger gains under increased semantic specificity.

% By leveraging retrieved image exemplars, our method improves semantic localization in both common and fine-grained settings without requiring model retraining. Across multiple encoders and datasets, we demonstrate consistent gains in object grounding, with improvements scaling under increased semantic specificity.

In online phase navigation, we show that these localization gains translate reliably in general-category regimes (OVON-syn), while long-tail settings (HSSD-rare) expose additional pipeline bottlenecks. Through failure analysis and controlled detector intervention, we identify fine-grained object detection as a key limiting factor in translating grounding improvements into navigation success. When object detection is improved, substantial additional gains are realized. These findings highlight that advances in queryable-map grounding and improvements in downstream mechanisms like object detection modules must evolve jointly to fully address long-tail embodied navigation.

% \vspace{-0.5cm}

\section*{Acknowledgment}

The authors thank the Department of Mathematics ``Tullio Levi-Civita'', University of Padua, and the CINECA/ISCRA initiative for providing access to computational resources. JRA acknowledges doctoral grant support from the University of Padua, Italy, and Fondazione Bruno Kessler, Italy.

%%%%%%%%%%%%%%%%%%%%%%%%%%%%%%%%%%%%%%%%%%%%%%%%%%%%%%%%%%%%%%%%%%%%%%%%%%%%%%%%


%%%%%%%%%%%%%%%%%%%%%%%%%%%%%%%%%%%%%%%%%%%%%%%%%%%%%%%%%%%%%%%%%%%%%%%%%%%%%%%%

% \nocite{*}
\bibliographystyle{IEEEtran}
\bibliography{egbib}

%%%%%%%%%%%%%%%%%%%%%%%%%%%%%%%%%%%%%%%%%%%%%%%%%%%%%%%%%%%%%%%%%%%%%%%%%%%%%%%%

% --------- APPENDIX --------

\section*{Appendix}
% \appendix

\subsection{HSSD-rare viewpoint generation pipeline}

\begin{figure}[!b]
    \centering
    \includegraphics[width=\columnwidth]{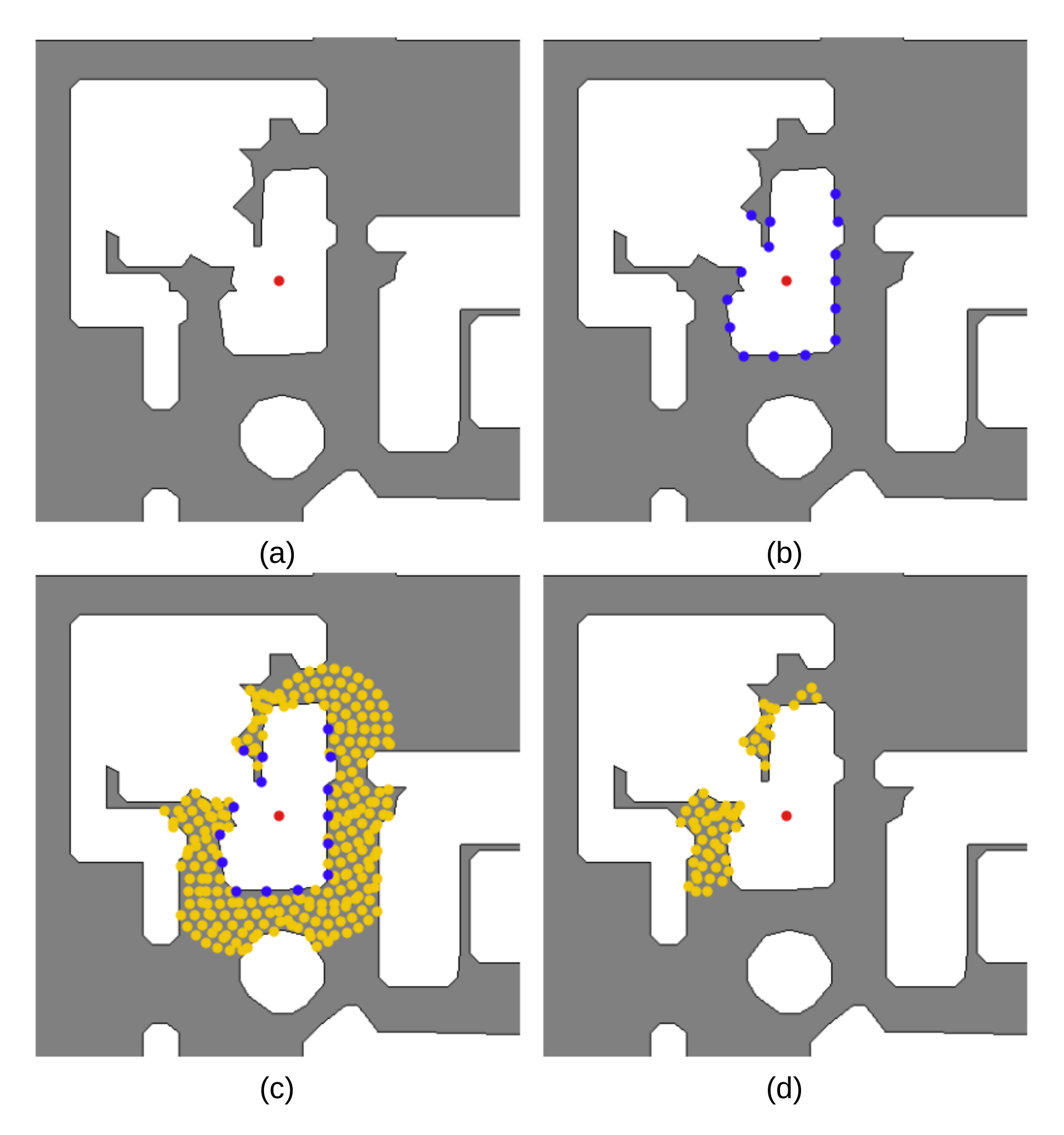}
    \caption{\textbf{Viewpoint-generation pipeline for a target object.}
    (a) The target object's location and dimensions are obtained, with its location shown in red.
    (b) A circle of radius $2\,\mathrm{m}$ is centered at the target, and points sampled at $0.5\,\mathrm{m}$ intervals along transitions from non-traversable to traversable space are retained as accessible boundary points, shown in blue.
    (c) Candidate viewpoints, shown in yellow, are generated as traversable points extending radially up to $1\,\mathrm{m}$ from each boundary point.
    (d) The candidates are validated using the procedure in Fig.~\ref{fig:hssd_rare_viewpoint_validate}, yielding the final viewpoint set.}
    \label{fig:hssd_rare_pipeline}
\end{figure}

To access the long-tail object categories in the HSSD dataset, we introduce a dedicated viewpoint-generation pipeline. HSSD does not provide category-specific viewpoints, nor does it include semantic bounding boxes that could be used to derive them directly. This pipeline therefore forms a core component of HSSD-rare. It uses the object positions and dimensions available in the HSSD metadata to generate reliable viewpoints for arbitrary target categories.

As shown in Fig.~\ref{fig:hssd_rare_pipeline}, the pipeline begins by retrieving the target object’s position and dimensions. Before generating viewpoints, we first account for the spatial constraints around the object, since obstacles and non-traversable regions determine how closely the agent can approach it. We consider a radial region of $2,\mathrm{m}$ around the target and evaluate points spaced at $0.5,\mathrm{m}$ intervals. Points that mark transitions from non-traversable to traversable space are retained as representative boundary points.

Candidate viewpoints are then generated radially from each boundary point, extending up to $1,\mathrm{m}$ while remaining within traversable space. This produces a set of accessible candidate locations around the target. Each candidate is subsequently validated to remove viewpoints from which the target is fully occluded.

\begin{figure}[!t]
    \centering
    \includegraphics[width=\columnwidth]{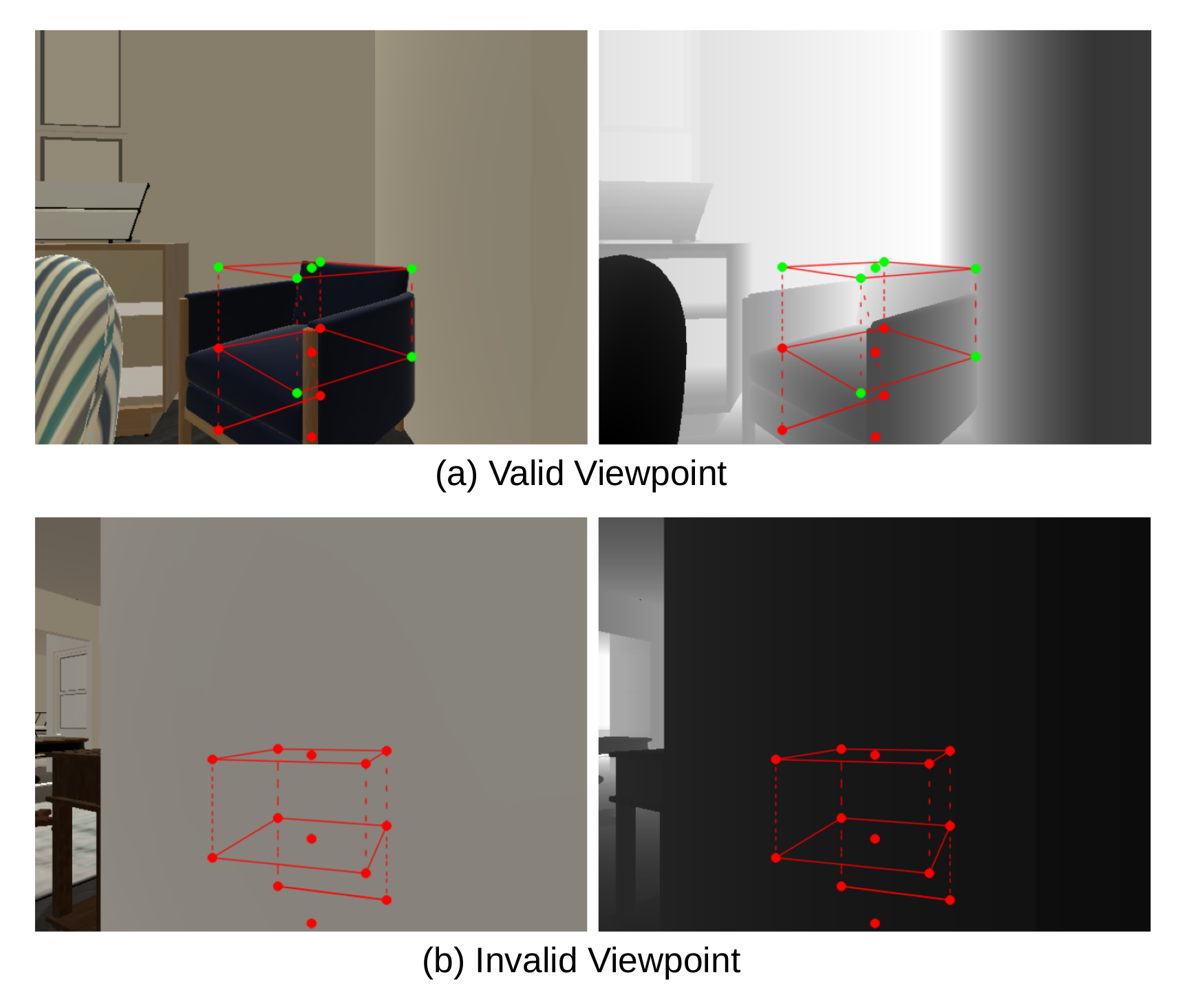}
    \caption{Examples of viewpoint validation using RGB and depth observations. The projected box approximates the target's 3D bounds in the observation frame, with green and red points indicating unobstructed and obstructed lines of sight, respectively. The connecting edges are shown only for visualization and are not used during validation. (a) A valid viewpoint, where at least one representative point is unobstructed. Some points are obstructed by nearer parts of the target itself. (b) An invalid viewpoint, where a wall occludes the target and all representative points are obstructed.}

\label{fig:hssd_rare_viewpoint_validate}
\end{figure}

Validation is based on whether the target is visible along the agent’s line of sight. For each candidate viewpoint, the agent is placed at the corresponding location and oriented toward the target. Using the target’s global position and dimensions, together with the agent’s position and orientation, we construct an approximate 3D bounding box at the expected target location and project it onto the agent’s 2D observation frame. The box is represented by three vertical planes, each defined by five representative points (as shown in Fig~\ref{fig:hssd_rare_viewpoint_validate}).

For each projected point, the expected distance from the agent is compared with the corresponding value in the depth image. If the observed depth is smaller than the expected distance, the line of sight to that point is considered obstructed. A viewpoint is therefore classified as invalid when all representative points are obstructed; otherwise, it is retained as valid.

\subsection{Semantic Coverage and Category Rarity}

\begin{table}[!b]
\centering
\setlength{\tabcolsep}{9pt}
\begin{tabular}{lcc}
\toprule
Dataset & \#Labels & $\text{OOV\%}_{(LVIS\cup COCO)}$ \\
\midrule
HSSD-rare   & 559 & 41.68\%  \\
OVON-HM3D   &   50 & 24.00\%  \\
\bottomrule
\end{tabular}
\caption{Ontology-anchored rarity analysis with LVIS$\cup$COCO.}
\label{tab:onto-lvis-coco}
\end{table}

A central objective of this work is to introduce a long-tail benchmark for ObjectGoal Navigation in Embodied AI. The breadth of HSSD-rare is first reflected in its scale: the dataset contains more than 500 object categories, compared with 50 in OVON-syn~\cite{yokoyama2024hm3d}, the long-tail subset of the OVON dataset. Although category count alone does not establish a long-tail distribution, it indicates a broader semantic coverage and motivates a more direct analysis of both semantic diversity and category rarity.

To quantify the semantic coverage of HSSD-rare, we compare its label set against the union of the \textbf{LVIS-train} and \textbf{COCO-80} ontologies and report the proportion of out-of-vocabulary (OOV) categories. A label is considered in-ontology if it matches a reference label exactly, shares its head noun, or achieves a fuzzy string similarity of at least $0.92$; otherwise, it is classified as OOV. This metric is intended to measure high-level semantic spread rather than fine-grained category specificity. In particular, the head-noun criterion allows detailed HSSD-rare labels to match broader LVIS or COCO categories, which makes the resulting OOV rate conservative with respect to fine-grained diversity. As shown in Table~\ref{tab:onto-lvis-coco}, HSSD-rare achieves an OOV rate of $41.68\%$, compared with $24.00\%$ for OVON-syn, providing quantitative support for its broader semantic coverage.

% To measure the fine-grained extent of the dataset, we adopt an LLM-based scoring protocol inspired by human judgments of object specificity~\cite{majumdar2024openeqa}. 

\begin{figure}[!t]
    \centering
    \includegraphics[width=0.50\textwidth]{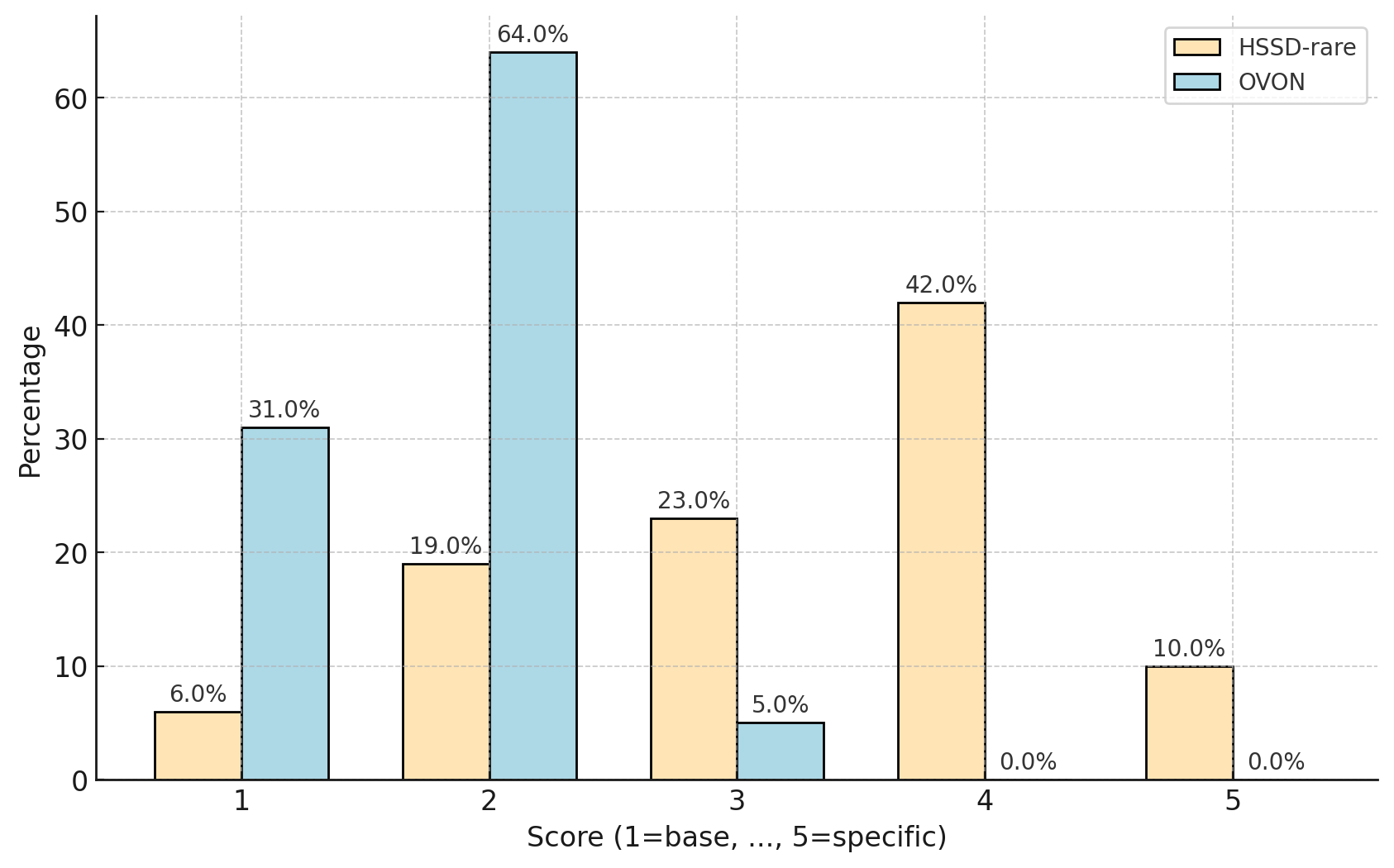}
    \caption{LLM-based rarity distribution (1--5 scale) for HSSD-rare vs.~OVON-syn, showing a clear shift toward more fine-grained categories in our dataset.}
    \label{fig:long_tail}
\end{figure}

To measure the fine-grained extent of the dataset, we adopt an LLM-based scoring protocol inspired by human judgments of object specificity. GPT-4.1 is prompted to assign each category a rarity score from 1 to 5. A score of 1 denotes a common base class, such as \emph{car}, \emph{dog}, or \emph{chair}; 2 denotes a common subtype or qualified base class, such as \emph{red car} or \emph{office chair}; 3 denotes a moderately specific or less common category, such as \emph{rocking chair} or \emph{fire truck}; 4 denotes a specific named subtype, model, brand, or geographically qualified class, such as \emph{IKEA bookshelf} or \emph{London taxi}; and 5 denotes a highly specific named entity or instance, such as \emph{IKEA BILLY bookcase} or \emph{1978 Citroën Méhari}. As shown in Fig.~\ref{fig:long_tail}, HSSD-rare contains a substantially larger concentration of categories at rarity levels 4--5 than OVON-syn, demonstrating its stronger long-tail character.

\subsection{Real-world experiment}
We deploy \approach\ on a LoCoBot wx250s robot in an active scenario. The robot is equipped with a single front-facing Intel RealSense D435 camera for RGB and depth input, and uses Hector SLAM for odometry. We experimented on 5 different episodes (\textit{“Red office chair”}, \textit{“Amazon Basics backpack”}, \textit{“Yellow Technogym bottle”}, \textit{“Android mascot toy”} and \textit{“Ikea Pong chair”}) and use a path planner followed by a path-tracking controller to reach the predicted location via velocity commands.
Since we only utilize the BLIP-2~\cite{li2023blip} backbone encoder, the whole pipeline requires less than 4GB of VRAM.
Our model using only textual queries was able to find just one out of five objects (\textit{“Red office chair”}), whereas both the model with web-sourced image queries and the one combining text and images successfully localized and reached four out of five objects, missing only the\textit{"Android mascot toy”}. The latter was likely not recognized properly due to its very small size, which makes it difficult for the feature extractor—operating at a global image level—to detect it accurately.

\begin{figure}[!t]
    \centering
    \includegraphics[width=0.3\textwidth]{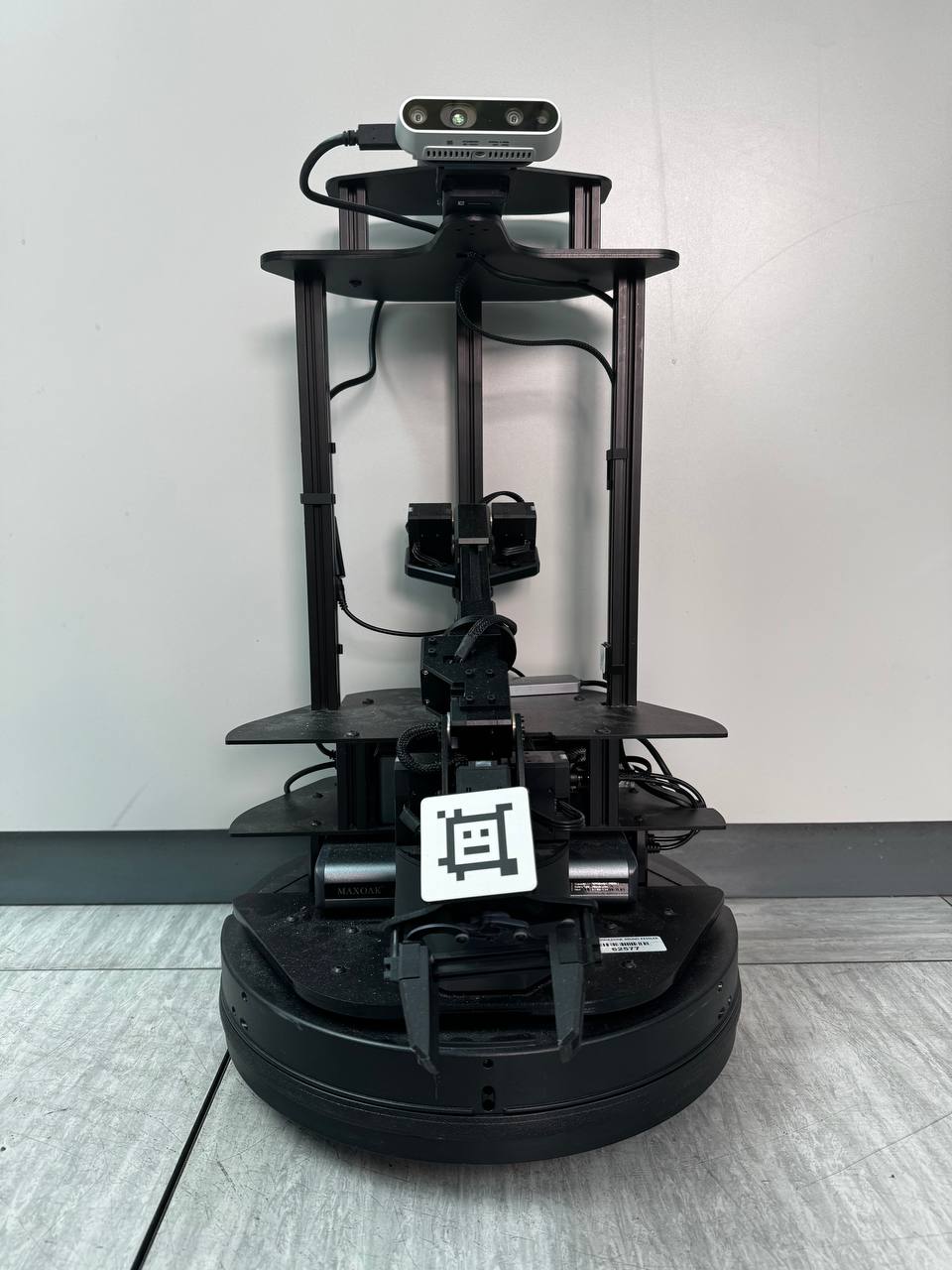}
    \caption{We performed a real-world experiment using a LoCoBot wx250s, utilizing both the installed RGB and depth cameras.}
    \label{fig:real_world_robot}
\end{figure}

\end{document}